\documentclass[runningheads]{llncs}

\usepackage{graphicx}
\usepackage{booktabs}
\usepackage{multirow}
\usepackage{subfig}
\usepackage{amsmath,amssymb}
\usepackage{array}
\usepackage{url}
\usepackage{xcolor}
\usepackage{float}
\usepackage{adjustbox}

\newcommand{\maybeincludegraphics}[2][]{%
\IfFileExists{#2}{\includegraphics[#1]{#2}}{%
\fbox{\begin{minipage}[c][0.13\textheight][c]{0.88\linewidth}\centering Figure placeholder\\\texttt{\detokenize{#2}}\end{minipage}}}}

\begin{document}

\title{The Personalization Gap in Wearable Stress Detection: Subject-Dependent and Cross-Modal Classification on WESAD}
\titlerunning{Personalization in Cross-Modal Stress Classification}

\author{Eric Oliver \inst{1} \and
Sagnik Dakshit\inst{2} }

\authorrunning{E.Oliver et al.}

\institute{The University of Texas at Tyler, Tyler, TX, USA\\
\email{}
\and
Kennesaw State University, Kennesaw, GA, USA \\
\email{sdakshit@kennesaw.edu}
}

\maketitle

\begin{abstract}
Wearable stress detection using physiological signals is strongly affected by personalization because stress responses vary across individuals. This paper revisits three-class stress classification on WESAD using a lightweight transformer trained directly on raw chest-worn physiological signals. We compare leave-one-subject-out cross-validation (LOSO-CV) with a mixed-subject, chronologically ordered subject-dependent protocol using the same architecture and modalities. LOSO-CV accuracy ranges from 55.00\% to 83.00\%, while the subject-dependent protocol reaches 99.73\% to 99.95\%. We interpret this gap not as evidence that stress detection is solved, but as a quantification of how strongly WESAD performance changes when the test subject is represented during training, while noting that WESAD's single-session structure does not isolate personalization from temporal and session-specific effects. The mean accuracy gain is 27.77 percentage points. We also evaluate subject-dependent cross-modal transfer by training on one modality and testing on another. ECG, EDA, respiration, temperature, and EMG show strong mutual classification compatibility under this protocol, with many macro F1-scores around 0.92 to 0.93, whereas accelerometer channels transfer poorly. PCA projections and in-class embedding variance provide qualitative context for this modality difference. Overall, the study uses WESAD to examine personalization and sensor compatibility while distinguishing subject-dependent results from true generalization.

\end{abstract}

\section{Introduction}

Wearable physiological sensing has made continuous and non-invasive stress monitoring increasingly feasible. Chest and wrist devices can record electrocardiogram (ECG), electrodermal activity (EDA), respiration, temperature, electromyography (EMG), and acceleration, creating multimodal data streams that can be modeled directly with modern deep learning methods. Stress monitoring is especially relevant for pervasive health because stress affects well-being, sleep, productivity, and long-term health, while mobile and wearable systems can support earlier recognition and personalized interventions. 

Despite this promise, reported performance in wearable stress detection is often difficult to interpret because evaluation protocols answer different deployment questions. A model evaluated on a held-out participant estimates performance for a new user, whereas a model evaluated on temporally disjoint windows from the same participants estimates within-session subject-dependent performance after user-specific patterns have been observed. These settings are both useful, but they should not be treated as equivalent. In particular, high performance under a mixed-subject split may reflect personalization, within-session separability, or subject-specific physiological structure rather than generalization to unseen individuals. A central difficulty is that stress physiology is highly individual. Baseline heart activity, skin conductance, breathing patterns, muscle activity, movement, and sensor placement can vary across people. Therefore, a stress classifier should be interpreted differently depending on whether it is tested on unseen users or on temporally disjoint data from users already represented during training. Leave-one-subject-out cross-validation (LOSO-CV) evaluates the former setting, while mixed-subject subject-dependent splits evaluate performance when subject identity is shared across training and testing. Cross-modal transfer adds a second practical question: whether a model trained on one physiological modality remains useful when one sensor fails but other sensors are available.

In this study, we investigate how much personalization matters for stress classification on WESAD. We use a lightweight transformer architecture trained directly on raw physiological signals and compare subject-independent LOSO-CV with a subject-dependent mixed-subject split. We further perform cross-modal testing under the subject-dependent protocol to examine classification compatibility across physiological signals. The paper does not claim that near-ceiling subject-dependent results establish stress detection that generalizes to unseen individuals. Instead, it treats the large difference between LOSO-CV and mixed-subject performance as the main empirical observation. The main contributions of this paper can be summarized as follows:

\begin{itemize}
    \item We compare subject-independent and subject-dependent stress classification on WESAD using the same lightweight transformer architecture to quantify the observed personalization gap. The mixed-subject protocol improves accuracy over LOSO-CV by an average of $27.77$ percentage points across ECG, EDA, respiration, temperature, EMG, and three ACC channels.
    \item To the best of author's knowledge, this study presents a subject-dependent cross-modal stress classification analysis in which models trained on one physiological modality are evaluated on the remaining modalities without changing the model architecture.
    \item We interpret cross-modal behavior using PCA projections and in-class embedding variance as qualitative context for the observed differences between physiological and accelerometer channels.
\end{itemize}

\section{Related Work}

Stress detection from physiological signals has been studied with both traditional machine learning and deep learning. WESAD was introduced by Schmidt et al. as a multimodal benchmark for wearable stress and affect detection using chest and wrist sensors \cite{schmidt2018introducing}. Subsequent work has used handcrafted features, multimodal fusion, and classifiers such as support vector machines, random forests, gradient boosting, and neural networks \cite{garg2021stress,mazumdarml,narwat2024multi}. More recent studies have explored CNNs, LSTMs, hybrid architectures, time-frequency representations, and transformer-based representation learning for physiological stress or affect recognition \cite{kumar2024deep,song2024stress,vaswani2017attention,wu2023transformer,gotz2023self,tang2024hierarchical}.

Direct comparison across WESAD studies is difficult because reported performance depends strongly on preprocessing, windowing, label definitions, modalities, and data splits. This is particularly important for personalization. Li et al. \cite{li2024comparison} compared personalized and generalized approaches for wearable emotion recognition, while domain generalization and transfer learning studies show that physiological recognition remains challenging when the test distribution differs from training \cite{albaladejo2023evaluating,fang2022towards}. Our contribution is complementary: using one lightweight transformer architecture, we explicitly quantify the performance gap between LOSO-CV and mixed-subject subject-dependent evaluation, and then examine subject-dependent cross-modal transfer. This focus is relevant for pervasive health systems because real wearable deployments face noisy signals, missing channels, sensor dropout, and user-specific baselines. Multimodal fusion assumes simultaneous sensor availability, whereas cross-modal testing asks whether a learned representation from one available modality can support classification on another. By holding subject identity constant while changing modality, our cross-modal analysis isolates modality compatibility within a personalized setting.

\section{Background and Methodology}

In this section, we differentiate between the concept of generalization and personalization to illustrate the central distinction of this paper followed by an illustration of the evaluation protocols and introduce our dataset and transformer architecture.

\subsection{Generalization and Personalization}

Generalization refers to performance on data not represented in training. In physiological stress detection, subject-independent generalization means that a model trained on several people can classify stress states for a new person. LOSO-CV is a strict protocol for this setting because one participant is held out for testing in each fold.

Personalization on the other hand refers to performance when the model has observed data from the same individual. In a subject-dependent protocol, training and test windows are disjoint but may come from the same participants. This setting is practically relevant for wearables that repeatedly monitor a user, but in a single-session dataset it may also retain temporal and session-specific structure and should not be interpreted as evidence of unseen-user or future-session generalization. This distinction is central: LOSO-CV asks whether a model transfers to a new person, while a mixed-subject split asks whether stress states are separable after the model has seen user-specific physiological structure.

\subsection{Evaluation Protocols}

We evaluate two protocols for within-modality classification which can be illustrated as:

\begin{itemize}
    \item \textbf{Subject-independent generalization:} LOSO-CV is used to test performance on unseen subjects. In each fold, one participant is held out for testing and the model is trained on the remaining participants.
    \item \textbf{Subject-dependent personalization:} A mixed-subject represents a model is trained and tested on different data windows from the same individuals, so it can learn each person’s physiological baseline and stress-response pattern. All subjects appear in both training and test sets, but the windows themselves are disjoint.
\end{itemize}

The mixed-subject protocol is intentionally subject-dependent. It is used to measure within-session subject-dependent classification rather than unseen-subject or future-session generalization. Because WESAD contains a single recording session per participant, this protocol cannot fully separate personalization from temporal autocorrelation or session-specific structure. Cross-modal experiments are restricted to this subject-dependent setting avoiding combining two sources of shift at once, namely subject shift and modality shift, where the baseline performance on subject shift alone is not strong. It also matches the main purpose of the paper, which is to examine cross-modal compatibility when subject identity is not excluded from training.

\subsection{Dataset}
\label{dataset}

This study uses the Wearable Stress and Affect Detection (WESAD) dataset \cite{schmidt2018introducing}. We focus on raw physiological signals collected from the chest-worn RespiBAN Professional device. WESAD contains recordings from 15 participants aged between 24 and 35 years who completed controlled experimental conditions designed to elicit affective states. This paper uses the three-class subset with labels corresponding to neutral, stress, and amusement. The analysis uses ECG, EDA, EMG, respiration (RESP), body temperature (TEMP), and three-axis accelerometer signals. The RespiBAN signals are sampled at a uniform rate of 700 Hz. For ACC, each axis is treated as an independent univariate channel, resulting in ACC-C1, ACC-C2, and ACC-C3. This design allows the same transformer architecture, patch embedding scheme, and positional encoding to be applied across all modalities without changing input dimensionality or adding modality-specific feature engineering. Raw signals were segmented into fixed-length windows of 700 samples, corresponding to 1 second at 700 Hz. To prevent normalization leakage, data partitioning was performed before normalization. In the subject-dependent protocol, windows were first assigned chronologically within each subject and modality to training, validation, and test partitions. Thus, temporally adjacent windows can remain across partition boundaries even though the windows themselves are disjoint. Z-score normalization parameters were then estimated using only the training partition for that subject and modality, and the resulting mean and standard deviation were applied unchanged to the validation and test partitions. Thus, validation and test windows did not influence normalization, early stopping, or model training. In the LOSO-CV protocol, normalization parameters were estimated from the training subjects in each fold and then applied to the held-out subject. No handcrafted features were extracted. This choice was made to evaluate whether a lightweight transformer can learn directly from raw physiological time-series windows.

\subsection{Transformer Architecture}
\label{archi}

\begin{figure}[t]
    \centering
    \maybeincludegraphics[width=0.82\textwidth]{Images/Screenshot 2025-12-14 201751.png}
    \caption{Overview of the lightweight transformer architecture used for raw 1D physiological signal classification. The same architecture is applied to ECG, EDA, respiration, temperature, EMG, and each ACC channel.}
    \label{fig:image7}
\end{figure}

The transformer architecture was originally introduced for sequence modeling using self-attention \cite{vaswani2017attention}. Although first developed for natural language processing, transformers have been adapted to 1D time-series data because they can capture temporal dependencies and relationships across sequence segments, similar to language. In this study, the same transformer structure is used for every physiological modality.

Each input window contains $700$ samples. The signal is divided into non-overlapping patches of $16$ samples, yielding $43$ patches per window. Each patch is linearly projected into a 64-dimensional embedding space, and a learned positional encoding is added to preserve temporal order. The transformer encoder contains three attention blocks. Each block uses multi-head self-attention with two attention heads, dropout with rate $0.1$, layer normalization with $\epsilon=10^{-6}$, and a feed-forward dense layer with ReLU activation. Residual connections are used throughout the attention blocks. After the transformer encoder, the representation is passed through a dense classification head with layer widths of 256, 256, 256, 128, 64, 32, and 16 units. A final dropout layer precedes a softmax output over the three classes. All models were trained with Adam using a learning rate of $0.001$, categorical cross-entropy loss, batch size of 32, and a maximum of 50 epochs. Early stopping with patience of 10 epochs was applied using a held-out validation subset from the training pool. Reported metrics in section \ref{results} correspond to the held-out test split for each protocol.

\section{Experimental Design and Results}
\label{results}

In this study, we perform and report results on three experiments:
\begin{itemize}
    \item Experiment A evaluates subject-independent within-modality performance using LOSO-CV.
    \item Experiment B evaluates subject-dependent within-modality performance using the mixed-subject temporal split.
    \item Experiment C uses the models from Experiment B without retraining and tests each model on the remaining modalities under the same subject-dependent setting for cross-modal performance evaluation. Target modalities use the corresponding training-partition normalization statistics described in Section~\ref{dataset}; validation and test statistics are not used. 
\end{itemize}  

The key interpretation is not only the absolute performance of each experiment, but also the change in performance between Experiment A and Experiment B.

\begin{table}[t]
\caption{Evaluation protocols and supported interpretations.}
\label{tab:protocol_claims}
\centering
\footnotesize
\begin{adjustbox}{max width=\textwidth}
\begin{tabular}{llll}
\toprule
\textbf{Exp.} & \textbf{Protocol} & \textbf{Subject overlap?} & \textbf{Supported interpretation} \\
\midrule
A & LOSO-CV within modality & No & Generalization to unseen users \\
B & 80:20 chronological mixed-subject split & Yes & Subject-dependent classification \\
C & Train on one modality, test on another & Yes & Subject-dependent modality compatibility \\
\bottomrule
\end{tabular}
\end{adjustbox}
\end{table}

\subsection{Experiment A: Subject-Independent Stress Classification}

\begin{table}[t]
\centering
\caption{Within-modality stress classification under LOSO-CV. These results evaluate subject-independent generalization to unseen participants.}
\label{tab:LOSOCV}
\footnotesize
\begin{tabular}{lccc}
\toprule
\textbf{Modality} & \textbf{Accuracy} & \textbf{Precision} & \textbf{Recall} \\
\midrule
ECG & \textbf{0.8300} & 0.5700 & \textbf{0.6630} \\
EDA & 0.7530 & 0.5400 & 0.5800 \\
Resp & 0.6900 & 0.5700 & 0.5400 \\
Temp & 0.5500 & 0.3920 & 0.3410 \\
EMG & 0.6067 & 0.5130 & 0.4400 \\
ACC-C1 & 0.7900 & 0.5400 & 0.6300 \\
ACC-C2 & 0.7500 & \textbf{0.7000} & 0.5900 \\
ACC-C3 & 0.8000 & 0.5300 & 0.6400 \\
\bottomrule
\end{tabular}
\end{table}

In the context of this paper, Experiment A provides the reference point for generalization. Table~\ref{tab:LOSOCV} reports the LOSO-CV results. ECG achieves the highest accuracy at $0.8300$ and the highest recall at $0.6630$. ACC-C3 and ACC-C1 also perform relatively well in accuracy, reaching $0.8000$ and $0.7900$, respectively. EDA reaches $0.7530$ accuracy, while respiration reaches $0.6900$. Temperature performs lowest, with $0.5500$ accuracy, $0.3920$ precision, and $0.3410$ recall. EMG also shows limited subject-independent performance with $0.6067$ accuracy. These results show that stress classification across unseen subjects remains difficult. The physiological response to stress is not uniform across people. Baseline heart activity, skin conductance, respiration patterns, muscle activity, and movement differ across individuals. LOSO-CV captures this difficulty because the model must classify a participant whose physiological distribution was not observed during training. 

\subsection{Experiment B: Subject-Dependent Stress Classification}

\begin{table}[t]
\centering
\caption{Within-modality stress classification under the mixed-subject subject-dependent protocol. All subjects contribute temporally disjoint windows to both training and testing.}
\label{tab:individual}
\footnotesize
\begin{tabular}{lccc}
\toprule
\textbf{Modality} & \textbf{Accuracy} & \textbf{Precision} & \textbf{Recall} \\
\midrule
ECG & 0.9994 & 0.9995 & 0.9994 \\
EDA & \textbf{0.9995} & \textbf{0.9995} & \textbf{0.9995} \\
Resp & \textbf{0.9995} & \textbf{0.9995} & \textbf{0.9995} \\
Temp & 0.9993 & 0.9993 & 0.9993 \\
EMG & 0.9991 & 0.9991 & 0.9991 \\
ACC-C1 & 0.9986 & 0.9986 & 0.9986 \\
ACC-C2 & 0.9973 & 0.9973 & 0.9973 \\
ACC-C3 & 0.9983 & 0.9983 & 0.9982 \\
\bottomrule
\end{tabular}
\end{table}

Table~\ref{tab:individual} reports the subject-dependent results representing the task of personalization. All modalities reach near-ceiling performance. EDA and respiration achieve the highest accuracy, precision, and recall at $0.9995$. ECG reaches $0.9994$ accuracy, temperature reaches $0.9993$, EMG reaches $0.9991$, and the ACC channels range from $0.9973$ to $0.9986$. These results should be interpreted as within-session subject-dependent performance as the mixed-subject protocol does not contain overlapping windows between training and testing, but it does contain the same participants in both sets. Therefore, the model can benefit from subject-specific physiological and condition-related structure. At the same time, Experiment B is not a test of generalization to unseen users or future sessions. The near-ceiling values are best read as an estimate of how much easier WESAD stress classification becomes when the model is allowed to observe data from the same participants during training; they should not be attributed exclusively to personalization because temporal autocorrelation and session-specific effects may also contribute.

\begin{table}[t]
\centering
\caption{Accuracy gain from LOSO-CV to the subject-dependent mixed-subject protocol. The gain is computed directly from Tables~\ref{tab:LOSOCV} and~\ref{tab:individual}.}
\label{tab:gap}
\footnotesize
\begin{tabular}{lccc}
\toprule
\textbf{Modality} & \textbf{LOSO Acc.} & \textbf{Mixed Acc.} & \textbf{Gain} \\
\midrule
ECG & 0.8300 & 0.9994 & 0.1694 \\
EDA & 0.7530 & 0.9995 & 0.2465 \\
Resp & 0.6900 & 0.9995 & 0.3095 \\
Temp & 0.5500 & 0.9993 & 0.4493 \\
EMG & 0.6067 & 0.9991 & 0.3924 \\
ACC-C1 & 0.7900 & 0.9986 & 0.2086 \\
ACC-C2 & 0.7500 & 0.9973 & 0.2473 \\
ACC-C3 & 0.8000 & 0.9983 & 0.1983 \\
\midrule
Mean & 0.7212 & 0.9989 & 0.2777 \\
\bottomrule
\end{tabular}
\end{table}

\textbf{Personalization Gap:} Table~\ref{tab:gap} summarizes the central result of the paper. Using the same architecture and the same modalities, the average accuracy increases from $0.7212$ under LOSO-CV to $0.9989$ under the subject-dependent protocol. The mean gain $is 0.2777$, or $27.77$ percent. The largest gain occurs for temperature, where accuracy increases by $44.93$ percentage points, followed by EMG with a gain of $39.24$ percentage points and respiration with a gain of 30.95 percentage points. ECG has the smallest gain at $16.94$ percentage points because it already performs best under LOSO-CV. This gap is not a minor methodological detail. It shows that the evaluation protocol strongly determines the apparent difficulty of WESAD stress classification. If the task is framed as new-user stress detection, the LOSO-CV results are the appropriate evidence. If the task is framed as personalized monitoring, the mixed-subject results show that the model can perform very well after being exposed to the same users. Both findings are useful, but they answer different questions. Temperature shows the largest protocol-dependent gain, while ECG shows the smallest, in part because ECG already performs best under LOSO-CV. Determining whether these differences arise from inter-individual physiology, temporal structure, or other modality-specific characteristics requires dedicated analysis and remains future work.

\subsection{Experiment C: Subject-Dependent Cross-Modal Transfer}
\label{cross-modal}

\begin{table}[t]
\caption{Cross-modal macro F1-score matrix under the mixed-subject subject-dependent protocol. Diagonal within-modality results are excluded.}
\label{tab:cross-testing}
\centering
\scriptsize
\setlength{\tabcolsep}{3.2pt}
\begin{adjustbox}{max width=\textwidth}
\begin{tabular}{lcccccccc}
\toprule
\textbf{Source} & \textbf{ECG} & \textbf{EDA} & \textbf{Resp} & \textbf{Temp} & \textbf{EMG} & \textbf{ACC-C1} & \textbf{ACC-C2} & \textbf{ACC-C3} \\
\midrule
ECG    & --   & 0.92 & 0.93 & 0.92 & 0.92 & 0.53 & 0.53 & 0.53 \\
EDA    & 0.92 & --   & 0.93 & 0.93 & 0.93 & 0.55 & 0.52 & 0.55 \\
Resp   & 0.93 & 0.93 & --   & 0.93 & 0.93 & 0.47 & 0.50 & 0.49 \\
Temp   & 0.93 & 0.93 & 0.93 & --   & 0.93 & 0.44 & 0.46 & 0.47 \\
EMG    & 0.92 & 0.92 & 0.92 & 0.92 & --   & 0.37 & 0.51 & 0.45 \\
ACC-C1 & 0.22 & 0.22 & 0.21 & 0.21 & 0.22 & --   & 0.48 & 0.47 \\
ACC-C2 & 0.25 & 0.26 & 0.24 & 0.26 & 0.26 & 0.49 & --   & 0.49 \\
ACC-C3 & 0.26 & 0.25 & 0.25 & 0.26 & 0.26 & 0.48 & 0.49 & --   \\
\bottomrule
\end{tabular}
\end{adjustbox}
\end{table}

\begin{figure}[t]
    \centering
    \includegraphics[width=0.92\textwidth]{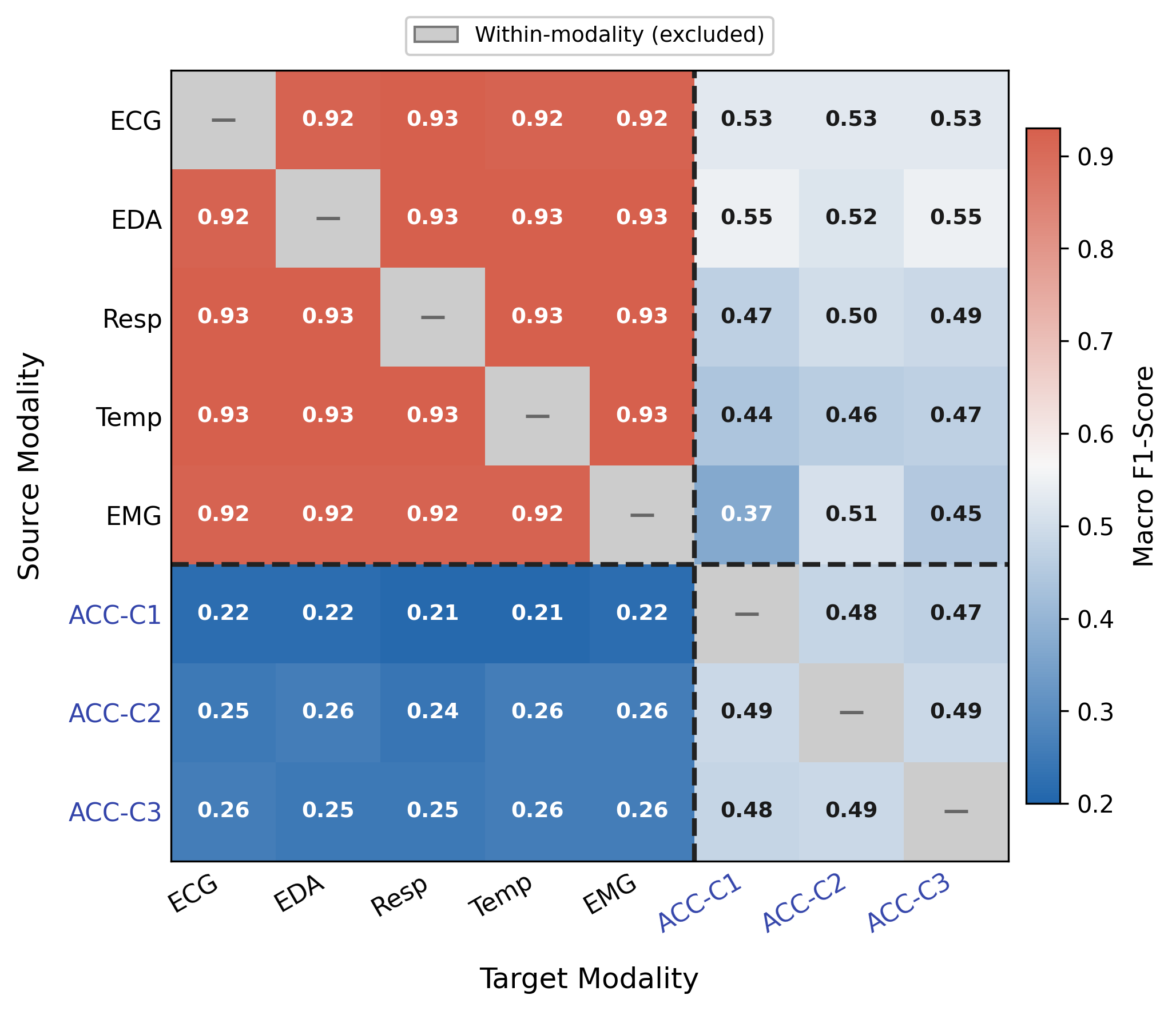}
    \caption{Cross-modal macro F1-score heatmap under the mixed-subject subject-dependent protocol. The autonomic physiological modalities show strong mutual compatibility, while ACC channels show weaker transfer.}
    \label{fig:crossmodal_heatmap}
\end{figure}

In this section we present and discuss our cross-modal analysis results as observed through experiment C as presented in Table~\ref{tab:cross-testing} and Figure \ref{fig:crossmodal_heatmap}. ECG, EDA, respiration, temperature, and EMG show strong transfer among one another. Many source-target pairs in this group reach F1-scores near $0.92$ to $0.93$. Respiration and temperature are particularly stable as target and source modalities, reaching approximately $0.93$ across several pairs. EDA and ECG also show strong compatibility with the other autonomic physiological signals. The accelerometer channels as expected behave differently as they are derived based on motion and are not direct physiological signals. When autonomic physiological models are tested on ACC targets, performance drops sharply. When ACC models are tested on ECG, EDA, respiration, temperature, or EMG, F1-scores are often around $0.21$ to $0.26$. Transfer among ACC channels is also limited, with F1-scores around $0.47$ to $0.49$. This suggests that ACC captures movement and context more than stress-specific autonomic physiology. It may still provide useful information in multimodal systems, but the present results do not support ACC as a strong standalone source for cross-modal stress transfer and begs the future uninvestigated question of are accelerometer signals a reliable medical signal for stress detection or more contextual support proxy. The results show that, once subject identity is shared between training and testing, some physiological modalities exhibit strong cross-modal classification compatibility under this synchronized protocol. They do not establish modality-invariant physiological representations or prove that a model trained on one modality will generalize to a new subject and a new modality at the same time. That harder problem remains open.

\section{Embedding-Based Interpretation}

Based on the results of Experiments B and C, this section examines the learned embedding spaces to better understand why some physiological modalities transfer more effectively than others. We use low-dimensional visualization and in-class variance analysis to relate representation compactness to the observed within-modality and cross-modal performance.

\subsection{Embedding Space Visualization}

The embedding space is a lower-dimensional representation where high-dimensional physiological windows are mapped to learned vectors. If the model learns a useful representation, windows from the same class should form compact regions, while different classes should be more separated. Principal component analysis (PCA) is commonly used to project high-dimensional representations in a lower-dimensional space \cite{mackiewicz1993principal} to use them as qualitative visualizations of class organization rather than as evidence of causal separability.

\begin{figure}[t]
\centering
\subfloat[ECG]{\maybeincludegraphics[width=0.31\textwidth]{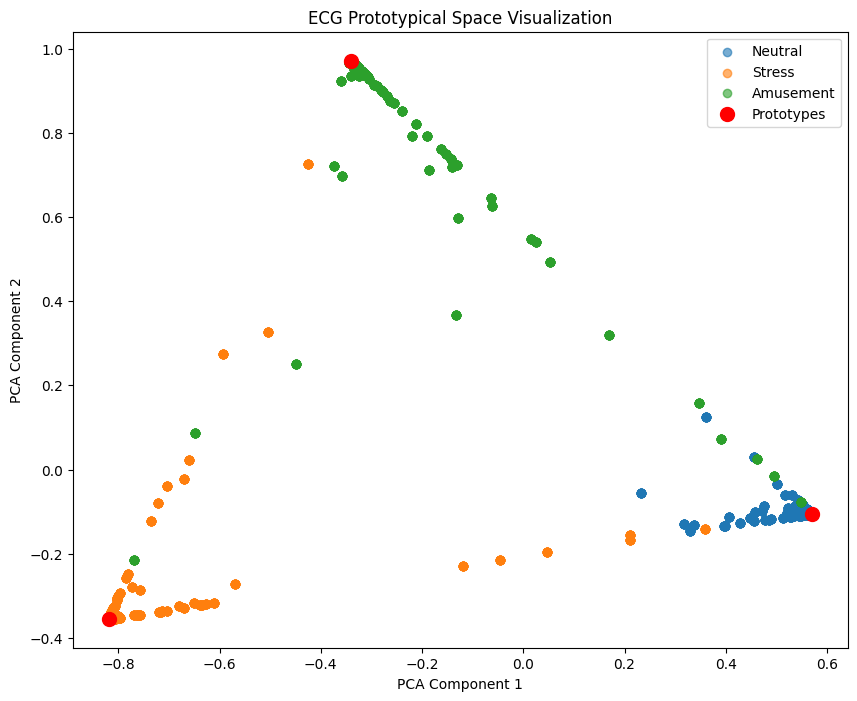}}
\hfill
\subfloat[EDA]{\maybeincludegraphics[width=0.31\textwidth]{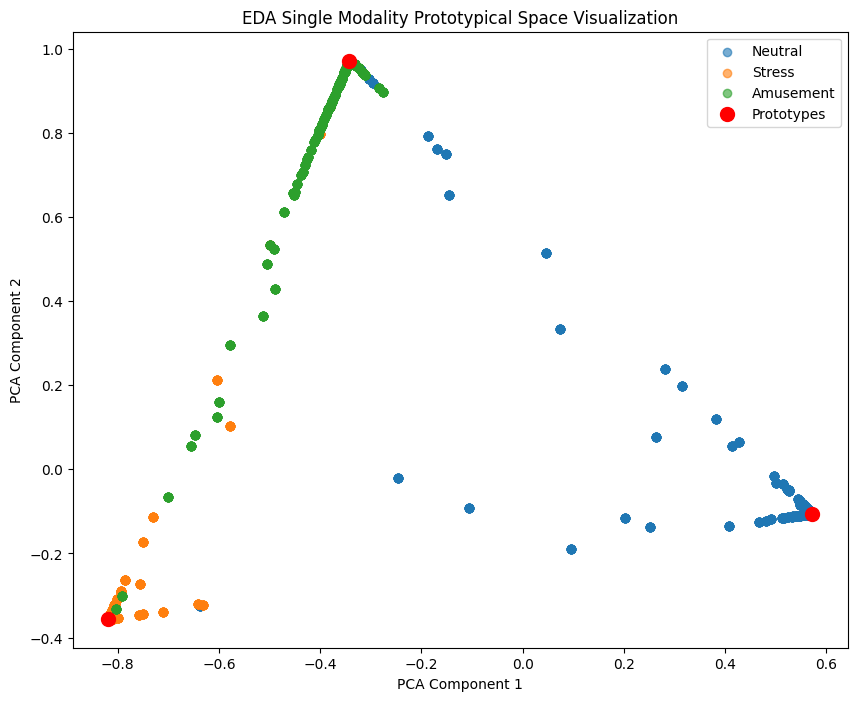}}
\hfill
\subfloat[Temp]{\maybeincludegraphics[width=0.31\textwidth]{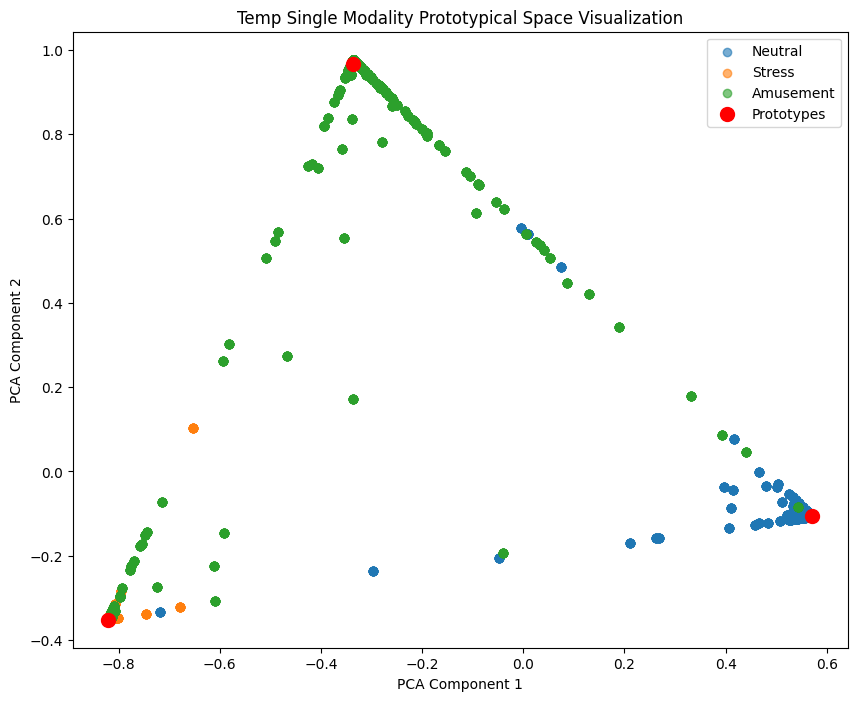}}

\subfloat[Resp]{\maybeincludegraphics[width=0.31\textwidth]{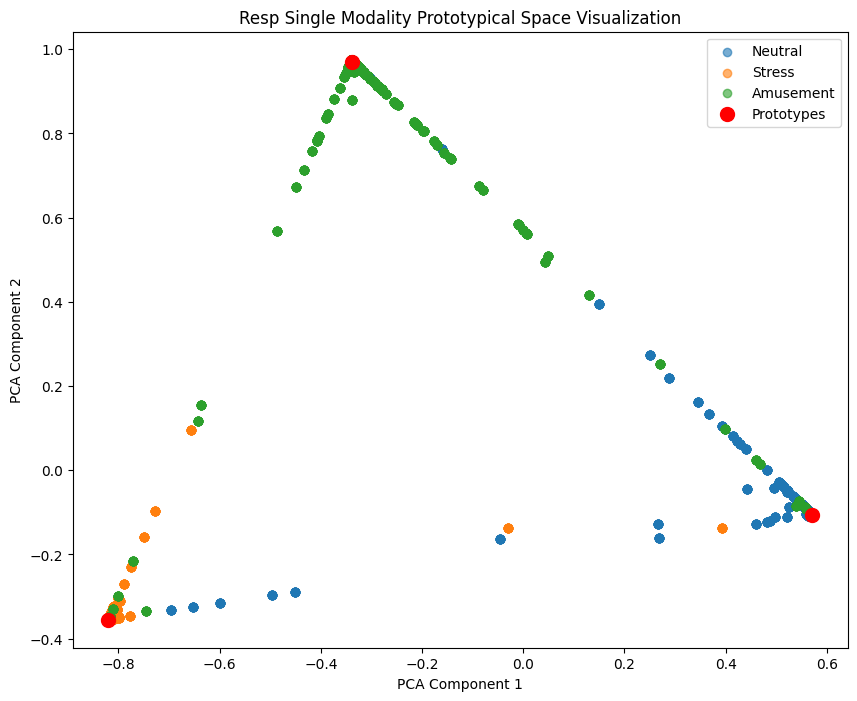}}
\hfill
\subfloat[EMG]{\maybeincludegraphics[width=0.31\textwidth]{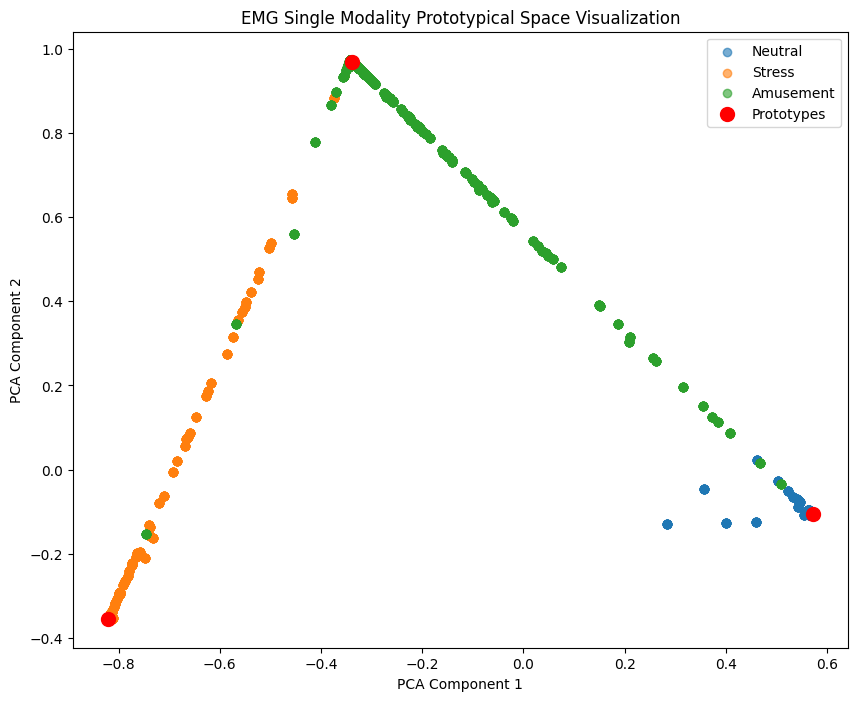}}
\hfill
\subfloat[ACC-C1]{\maybeincludegraphics[width=0.31\textwidth]{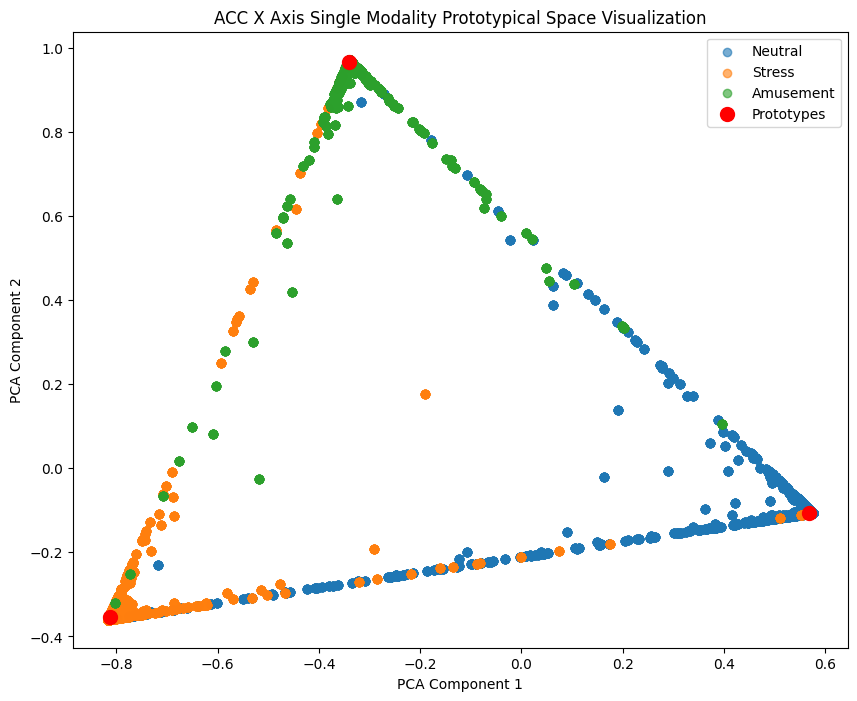}}
\caption{PCA projections of learned embedding spaces for selected single-modality transformer models. ACC-C1 is shown as the representative accelerometer channel.}
\label{fig:pca}
\end{figure}

Figure~\ref{fig:pca} shows the PCA projections of learned embeddings from the subject-dependent models of Experiment B. ECG and EDA show clearer class organization over other modalities, which is consistent with their strong within-modality and cross-modal performance. Respiration and temperature show useful separation, but also significant overlap between the classifying states. EMG and ACC-C1 show a greater overlap, especially between neutral and amusement. This visual pattern matches the quantitative results, where modalities with compact and separable embeddings tend to transfer better than modalities whose embeddings are more dispersed.

\subsection{In-Class Variance}

\begin{table}[H]
\centering
\caption{In-class variance of learned embeddings for neutral, stress, and amusement. Lower values indicate more compact class representations.}
\label{tab:variance}
\footnotesize
\begin{tabular}{lccc}
\toprule
\textbf{Modality} & \textbf{Neutral} & \textbf{Stress} & \textbf{Amusement} \\
\midrule
ECG & 3.044E-05 & 5.402E-04 & 2.728E-03 \\
EDA & 5.117E-04 & 3.889E-04 & 1.078E-03 \\
RESP & 3.994E-04 & 1.914E-04 & 2.355E-03 \\
TEMP & 2.199E-04 & 1.080E-04 & 7.786E-03 \\
EMG & 3.332E-06 & 5.116E-04 & 3.214E-03 \\
ACC-C1 & 1.493E-03 & 3.525E-03 & 8.287E-04 \\
ACC-C2 & 1.182E-02 & 3.001E-03 & 4.098E-03 \\
ACC-C3 & 1.984E-03 & 1.946E-03 & 5.395E-03 \\
\bottomrule
\end{tabular}
\end{table}

In Table~\ref{tab:variance}, we report the in-class variance of the learned embeddings for neutral, stress, and amusement classes. These values provide qualitative context for the cross-modal patterns. ECG, EDA, and respiration generally have low variance for neutral and stress classes, and they also show strong cross-modal transfer. Stress responses are closely tied to autonomic activity, so these signals tend to encode related physiological changes. Temperature and EMG also transfer well among the autonomic group in Experiment C, but their amusement variance is higher, which may explain some class-level ambiguity observed in the PCA visualizations in Figure \ref{fig:pca}. ACC channels have larger variance values overall, especially ACC-C2 in the neutral class and ACC-C3 in the amusement class. This is qualitatively consistent with movement-based signals being less stable for stress representation in this setup.

The variance analysis does not prove a causal relationship between embedding compactness and transferability and is not evidence of cross-modal representation alignment. It does, however, provide a useful exploratory framework. Lower in-class variance suggests that a modality creates a more stable class representation, which can make it easier for the same architecture to classify another modality in a subject-dependent setting. Higher variance suggests that the learned representation is more sensitive to motion artifacts, posture, or context, which can reduce transferability.

\section{Comparison with Existing WESAD Studies}
\label{comparison}

Existing WESAD studies report a wide range of results because they use different sensors, preprocessing pipelines, features, model architectures, class definitions, and data splits. Classical machine learning studies have reported moderate performance for three-class stress classification, while more recent deep learning and ensemble approaches often report higher scores \cite{garg2021stress,schmidt2018introducing,kumar2024resp,li2024comparison,narwat2024multi,mazumdarml}. Direct comparison is therefore difficult unless the evaluation protocol is the same and redundant as the scope of this paper does not include developing state-of-the-art classifier or benchmarking. The present study is distinct from most prior work in two ways. First, it explicitly compares LOSO-CV and subject-dependent mixed-subject evaluation using the same transformer architecture. This comparison shows that protocol choice has a large effect on reported performance. Second, it examines cross-modal transfer under a subject-dependent setting, which is less commonly reported than multimodal fusion. Therefore, the main value of the paper is not a claim of state-of-the-art stress detection. The value is the analysis of how personalization changes performance and how modality transfer behaves when subject identity is shared between training and testing.

\section{Limitations and Discussion}

In this section, we acknowledge the limitations that will also guide future work. First, WESAD is a controlled laboratory dataset, so future studies should test whether the observed personalization gap persists in longer, noisier, and more naturalistic recordings. Second, the present analysis uses one transformer architecture to keep the comparison focused on protocol and modality effects. Future work will compare CNN, LSTM, transformer, and hybrid models under identical subject-independent, subject-dependent, and cross-modal protocols.

Although the normalization procedure avoids validation/test statistics entering normalization, the subject-dependent results should still be interpreted cautiously because WESAD contains single-session laboratory recordings. One-second windows from the same recording may contain temporal autocorrelation and subject/session-specific structure, so the raw number of windows should not be treated as an equivalent number of independent observations for uncertainty estimation. For this reason, the mixed-subject protocol is best understood as evaluating within-session subject-dependent classification, not generalization to future sessions, future days, or unseen users.

The results point to three main observations:
\begin{itemize}
    \item First, subject-independent stress classification remains challenging on WESAD. The LOSO-CV results range from $0.5500$ to $0.8300$ accuracy, showing that the same architecture does not learn a fully transferable stress representation across people. This is consistent with the physiological variability expected in stress responses.
    \item Second, the subject-dependent protocol has a large effect. Under the mixed-subject protocol, the same model reaches near-ceiling performance across all modalities. This does not mean that stress detection has been solved in a general setting. It means that, within WESAD's controlled recording structure, the model can classify temporally disjoint windows from users whose data were also present during training, with personalization, temporal autocorrelation, and session structure potentially contributing to the observed gap. For new-user deployment, LOSO-CV remains the more relevant evaluation. 
    \item Third, cross-modal transfer is much stronger among autonomic physiological signals than among accelerometer channels. ECG, EDA, respiration, temperature, and EMG are all linked to stress-related physiological processes, although through different mechanisms. Their high cross-modal scores show classification compatibility across these synchronized signals in the subject-dependent setting, but do not by themselves establish a shared physiological representation. ACC channels capture movement, which may correlate with affective state in some contexts but is not as directly tied to autonomic stress physiology. 
\end{itemize} 

This study also raises a broader methodological point: when WESAD is split subject-dependently, very high performance can be obtained, making the setting useful for studying personalization, but it should be labeled explicitly as such. If a paper reports near-ceiling mixed-subject performance without separating it from subject-independent generalization, the result can be misleading. The personalization gap reported here provides a concrete way to communicate this difference.

\section{Conclusion}

This paper reframes stress classification on WESAD around a simple question: how much does personalization matter? Using the same lightweight transformer architecture, subject-independent LOSO-CV produced accuracies between $55.00\%$ and $83.00\%$, while a subject-dependent mixed-subject protocol produced near-ceiling accuracies between $99.73\%$ and $99.95\%$. The average accuracy gain was $27.77$ percentage points. This gap is the central finding of the paper. It shows that WESAD stress classification is strongly affected by whether the test subject is present in the training data. The cross-modal results further show that ECG, EDA, respiration, temperature, and EMG have strong subject-dependent classification compatibility, with many source-target pairs reaching F1-scores around $0.92$ to $0.93$. ACC channels transfer poorly under this protocol. PCA projections and in-class variance analysis provide qualitative context for these patterns but do not establish cross-modal representation alignment. Overall, the study supports personalized and sensor-flexible stress monitoring as a promising direction, while also emphasizing that subject-dependent results should not be confused with unseen-subject generalization. Future work should evaluate repeated-session personalization, independent datasets such as CASE and DEAP \cite{sharma2019dataset,koelstra2011deap}, and stronger controls for temporal and subject-specific structure.

\bibliographystyle{splncs04}
\bibliography{bib}

\end{document}